\documentclass{article} 
\usepackage{iclr2020_conference,times}

\usepackage{amsmath,amsfonts,bm}

\def\eqref#1{equation~\ref{#1}}

\def\1{\bm{1}}

\DeclareMathAlphabet{\mathsfit}{\encodingdefault}{\sfdefault}{m}{sl}
\SetMathAlphabet{\mathsfit}{bold}{\encodingdefault}{\sfdefault}{bx}{n}

\usepackage{accents}
\usepackage{hyperref}
\usepackage{url}
\usepackage{makecell}
\usepackage{amsfonts}       
\usepackage{nicefrac}       
\usepackage{microtype}      
\usepackage{color}
\usepackage{tabularx}
\usepackage[linesnumbered,ruled]{algorithm2e}

\SetCommentSty{mycommfont}
\usepackage{times}
\usepackage{epsfig}
\usepackage{graphicx}
\usepackage{amsmath}
\usepackage{amssymb}
\usepackage{enumitem}
\usepackage{graphicx}
\usepackage{xspace}
\usepackage{balance}
\usepackage{multirow}
\usepackage{mathtools}
\usepackage{rotating}
\usepackage{tabulary}
\usepackage{array}
\usepackage{subcaption}
\usepackage[export]{adjustbox}
\usepackage{ctable}
\usepackage{physics}
\usepackage[title]{appendix}
\usepackage{xcolor}

\newcommand{\ie}{\textit{i}.\textit{e}.}
\newcommand{\eg}{\textit{e}.\textit{g}.}

\newcolumntype{x}[1]{>{\centering\let\newline\\\arraybackslash\hspace{0pt}}p{#1}}

\graphicspath{{./figs/}}

\title{DivideMix: Learning with Noisy Labels as Semi-supervised Learning}

\iclrfinalcopy

\author{Junnan Li, Richard Socher, Steven C.H. Hoi\\
	Salesforce Research\\
	\texttt{\{junnan.li,rsocher,shoi\}@salesforce.com}
}

\begin{document}

\maketitle
\begin{abstract}
Deep neural networks are known to be annotation-hungry.
Numerous efforts have been devoted to reducing the annotation cost when learning with deep networks.
Two prominent directions include learning with noisy labels and semi-supervised learning by exploiting unlabeled data. In this work,
we propose $\mathrm{DivideMix}$,
a novel framework for learning with noisy labels by leveraging semi-supervised learning techniques.
In particular, $\mathrm{DivideMix}$ models the per-sample loss distribution with a mixture model to dynamically divide the training data into a labeled set with clean samples and an unlabeled set with noisy samples, 
and trains the model on both the labeled and unlabeled data in a semi-supervised manner.
To avoid confirmation bias,
we simultaneously train two diverged networks
where each network uses the dataset division from the other network.
During the semi-supervised training phase,
we improve the $\mathrm{MixMatch}$ strategy by performing label co-refinement and label co-guessing on labeled and unlabeled samples, respectively.
Experiments on multiple benchmark datasets demonstrate substantial improvements over state-of-the-art methods.
Code is available at \textcolor{magenta}{\url{https://github.com/LiJunnan1992/DivideMix}}.
\end{abstract} 
\section{Introduction}
\label{sec:introduction}
The remarkable success in training deep neural networks (DNNs) is largely attributed to the collection of large datasets with human annotated labels.
However, it is extremely expensive and time-consuming to label extensive data with high-quality annotations.
On the other hand,
there exist alternative and inexpensive methods for mining large-scale data with labels,
such as querying commercial search engines~\citep{webvision}, 
downloading social media images with tags~\citep{Mahajan_ECCV_2018},
leveraging machine-generated labels~\citep{OpenImages},
or using a single annotator to label each sample~\citep{Tanno_2019_CVPR}.
These alternative methods inevitably yield samples with \textit{noisy labels}.
A recent study~\citep{Zhang_ICLR_2017} shows that DNNs can easily overfit to noisy labels and results in poor generalization performance.

Existing methods on learning with noisy labels (LNL) primarily take a loss correction approach.
Some methods estimate the noise transition matrix and use it to correct the loss function~\citep{Giorgio_CVPR_2017,Goldberger_ICLR_2017}.
However, correctly estimating the noise transition matrix is challenging.
Some methods leverage the predictions from DNNs to correct labels and modify the loss accordingly~\citep{Reed_2015_ICLR,Tanaka_CVPR_2018}.
These methods do not perform well under high noise ratio as the predictions from DNNs would dominate training and cause overfitting.
To overcome this,
\cite{Arazo_ICML_2019} adopt $\mathrm{MixUp}$~\citep{mixup} augmentation.
Another approach selects or reweights samples so that noisy samples contribute less to the loss~\citep{Jiang_ICML_2018,Ren_ICML_2018}.
A challenging issue is to design a reliable criteria to select clean samples.
It has been shown that DNNs tend to learn simple patterns first before fitting label noise~\citep{Arpit_ICML_2017}.
Therefore,
many methods treat samples with small loss as clean ones~\citep{Jiang_ICML_2018,Arazo_ICML_2019}.
Among those methods,
Co-teaching~\citep{co-teaching} and Co-teaching$+$~\citep{Yu_ICML_2019} train two networks where each network selects small-loss samples in a mini-batch to train the other.

 
Another active area of research that also aims to reduce annotation cost is semi-supervised learning (SSL).
In SSL,
the training data consists of unlabeled samples in addition to the labeled samples.
Significant progress has been made in leveraging unlabeled samples by enforcing the model to produce low entropy predictions on unlabeled data~\citep{Grandvalet_NIPS_2005} or consistent predictions on perturbed input~\citep{Laine_ICLR_2017,Tarvainen_NIPS_17,VAT}.
Recently, \cite{mixmatch} propose $\mathrm{MixMatch}$,
which unifies several dominant SSL approaches in one framework and achieves state-of-the-art performance.

Despite the individual advances in LNL and SSL, their connection has been underexplored.
In this work, we propose $\mathrm{DivideMix}$, which addresses learning with label noise in a semi-supervised manner. Different from most existing LNL approaches, $\mathrm{DivideMix}$ discards the sample labels that are highly likely to be noisy, and leverages the noisy samples as unlabeled data to regularize the model from overfitting and improve generalization performance. The key contributions of this work are:
\vspace{-\topsep}
\begin{itemize}
	\setlength\itemsep{0pt}
	\item 	
	We propose co-divide,
	which trains two networks simultaneously.
	For each network, we dynamically fit a Gaussian Mixture Model (GMM) on its per-sample loss distribution to divide the training samples into a labeled set and an unlabeled set.
	The divided data is then used to train the \textit{other} network.
    Co-divide keeps the two networks diverged,
    so that they can filter different types of error and avoid confirmation bias in self-training.
	\item
	During SSL phase,
	we improve $\mathrm{MixMatch}$ with label co-refinement and co-guessing to account for label noise.
	For labeled samples,
	we refine their ground-truth labels using the network's predictions guided by the GMM for the other network.
	For unlabeled samples,
	we use the ensemble of both networks to make reliable guesses for their labels.
	\item 
	We experimentally show that $\mathrm{DivideMix}$ significantly advances state-of-the-art results on multiple benchmarks with different types and levels of label noise. 
	We also provide extensive ablation study and qualitative results to examine the effect of different components.
\end{itemize}
\vspace{-\topsep}

\section{Related Work}
\label{sec:literature}
\subsection{Learning with Noisy Labels}
Most existing methods for training DNNs with noisy labels seek to correct the loss function.
The correction can be categorized in two types.
The first type treats all samples equally and correct loss either explicitly or implicitly through relabeling the noisy samples.
For relabeling methods,
the noisy samples are modeled with directed graphical models~\citep{Tong_CVPR_2015},
Conditional Random Fields~\citep{Vahdat_NIPS_2017},
knowledge graph~\citep{Li_ICCV_17},
or DNNs~\citep{Andreas_CVPR_2017,Lee_CVPR_2018}.
However, 
they require access to a small set of clean samples.
Recently,
\cite{Tanaka_CVPR_2018} and \cite{Yi_2019_CVPR} propose iterative methods which relabel samples using network predictions.
For explicit loss correction.
\cite{Reed_2015_ICLR} propose a bootstrapping method which modifies the loss with model predictions,
and \cite{Ma_ICML_2018} improve the bootstrapping method by exploiting the dimensionality of feature subspaces.
\cite{Giorgio_CVPR_2017} estimate the label corruption matrix for loss correction,
and \cite{Hendrycks_NIPS_2018} improve the corruption matrix by using a clean set of data.
The second type of correction focuses on reweighting training samples or separating clean and noisy samples, which results in correcting the loss function~\citep{Abstention,Nikola_ICML_2019}.
A common method is to consider samples with smaller loss as clean ones~\citep{Shen_ICML_2019}. 
\cite{Jiang_ICML_2018} train a mentor network to guide a student network by assigning weights to samples.
\cite{Ren_ICML_2018} reweight samples based on their gradient directions.
\cite{Chen_ICML_2019} apply cross validation to identify clean samples. 
\cite{Arazo_ICML_2019} calculate sample weights by modeling per-sample loss with a mixture model.
\cite{co-teaching} train two networks which select small-loss samples within each mini-batch to train each other,
and \cite{Yu_ICML_2019} improve it by updating the network on disagreement data to keep the two networks diverged.

Contrary to all aforementioned methods,
our method discards the labels that are highly likely to be noisy,
and utilize the noisy samples as unlabeled data to regularize training in a SSL manner.
\cite{WACV} and \cite{Recycling} have shown that SSL method is effective in LNL.
However, their methods do not perform well under high levels of noise,
whereas our method can better distinguish and utilize noisy samples.
Besides leveraging SSL,
our method also introduces other advantages.
Compared to self-training methods~\citep{Jiang_ICML_2018,Arazo_ICML_2019},
our method can avoid the confirmation bias problem~\citep{Tarvainen_NIPS_17} by training two networks to filter error for each other.
Compared to Co-teaching~\citep{co-teaching} and Co-teaching$+$~\citep{Yu_ICML_2019},
our method is more robust to noise by enabling the two networks to teach each other implicitly at each epoch (co-divide) and explicitly at each mini-batch (label co-refinement and co-guessing).

\subsection{Semi-supervised Learning}
SSL methods aim to improve the model's performance by leveraging unlabeled data.
Current state-of-the-art SSL methods mostly involve adding an additional loss term on unlabeled data to regularize training.
The regularization falls into two classes:
consistency regularization~\citep{Laine_ICLR_2017,Tarvainen_NIPS_17,VAT} enforces the model to produce consistent predictions on augmented input data;
entropy minimization~\citep{Grandvalet_NIPS_2005,Pseudo} encourages the model to give high-confidence predictions on unlabeled data.
Recently,
\cite{mixmatch} propose $\mathrm{MixMatch}$,
which unifies consistency regularization, entropy minimization, and the $\mathrm{MixUp}$~\citep{mixup} regularization into one framework.

\section{Method}
\label{sec:method}

\begin{figure*}[!t]
 \centering

   \includegraphics[width=0.95\textwidth]{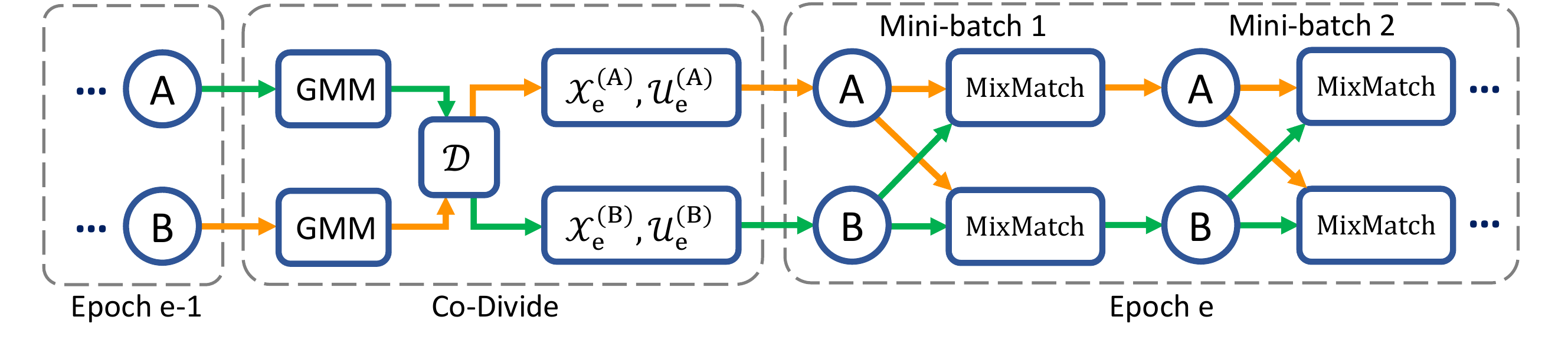}
   \vspace{-1.5ex}
   \caption
  	{
  	\small
		$\mathrm{DivideMix}$ trains two networks (A and B) simultaneously. At each epoch, a network models its per-sample loss distribution with a GMM to divide the dataset into a labeled set (mostly clean) and an unlabeled set (mostly noisy),
		which is then used as training data for the other network (\ie~co-divide).
		At each mini-batch,
		a network performs semi-supervised training using an improved $\mathrm{MixMatch}$ method.
		We perform label co-refinement on the labeled samples and label co-guessing on the unlabeled samples.
	 } 
 \vspace{-0.5ex}
  \label{fig:framework}
 \end{figure*}
\begin{algorithm}[!t]
	
	\DontPrintSemicolon
	\small
	\textbf{Input:} $\theta^{(1)}$ and $\theta^{(2)}$, training dataset $(\mathcal{X},\mathcal{Y})$, clean probability threshold  $\tau$, number of augmentations $M$, sharpening temperature $T$, unsupervised loss weight $\lambda_u$, $\mathrm{Beta}$ distribution parameter $\alpha$ for $\mathrm{MixMatch}$. \\

	$\theta^{(1)},\theta^{(2)}=\mathrm{WarmUp}(\mathcal{X},\mathcal{Y},\theta^{(1)},\theta^{(2)})$ \tcp*{standard training (with confidence penalty)}
	\While{$e<\mathrm{MaxEpoch}$}    
	{	
	$\mathcal{W}^{(2)} = \mathrm{GMM}(\mathcal{X},\mathcal{Y},\theta^{(1)})$ \tcp*{model per-sample loss with $\theta^{(1)}$ to obtain clean proabability for $\theta^{(2)}$}
	$\mathcal{W}^{(1)} = \mathrm{GMM}(\mathcal{X},\mathcal{Y},\theta^{(2)})$ \tcp*{model per-sample loss with $\theta^{(2)}$ to obtain clean proabability for $\theta^{(1)}$}

	\For( \tcp*[f]{train the two networks one by one}) {$k=1,2$}    
    {
	$\mathcal{X}^{(k)}_e = \{(x_i,y_i,w_i) | w_i\ge\tau, \forall (x_i,y_i,w_i) \in (\mathcal{X},\mathcal{Y},\mathcal{W}^{(k)})\}$\tcp*{labeled training set for $\theta^{(k)}$}
	$\mathcal{U}^{(k)}_e = \{x_i | w_i<\tau, \forall (x_i,w_i) \in (\mathcal{X},\mathcal{W}^{(k)})\}$  \tcp*{unlabeled training set for $\theta^{(k)}$}
    	
		\For {$\mathrm{iter}=1$ \KwTo $\mathrm{num\_iters}$}
		{
			From $\mathcal{X}^{(k)}_e$, draw a mini-batch $\{(x_b,y_b,w_b);b\in(1,...,B)\}$ \\
			From $\mathcal{U}^{(k)}_e$, draw a mini-batch $\{u_b;b\in(1,...,B)\}$ \\
			\For{$b=1$ \KwTo $B$}
			{
				\For{$m=1$ \KwTo $M$}
				{
					$\hat{x}_{b,m}=\mathrm{Augment}(x_b)$  \tcp*{apply $m^{th}$ round of augmentation to $x_b$}
					$\hat{u}_{b,m}=\mathrm{Augment}(u_b)$  \tcp*{apply $m^{th}$ round of augmentation to $u_b$}
				}
				${p}_{b}=\frac{1}{M}\sum_{m}\mathrm{p_{model}}(\hat{x}_{b,m};\theta^{(k)})$ \tcp*{average the predictions across augmentations of $x_b$}
				$\bar{y}_b=w_b y_b+(1-w_b){p}_{b}$ \\
				 	\tcp*{refine ground-truth label guided by the clean probability produced by the other network}
				$\hat{y}_b=\mathrm{Sharpen}(\bar{y}_b,T)$ 	\tcp*{apply temperature sharpening to the refined label}
				$\bar{q}_{b}=\frac{1}{2M}\sum_{m}\big(\mathrm{p_{model}}(\hat{u}_{b,m};\theta^{(1)})+\mathrm{p_{model}}(\hat{u}_{b,m};\theta^{(2)})\big)$\\
				 \tcp*{co-guessing: average the predictions from both networks across augmentations of $u_b$}
				${q}_{b}=\mathrm{Sharpen}(\bar{q}_{b},T)$		\tcp*{apply temperature sharpening to the guessed label}		    
			}
			$\hat{\mathcal{X}}=\{(\hat{x}_{b,m},\hat{y}_b);b\in(1,...,B),m\in(1,...,M)\}$	\tcp*{augmented labeled mini-batch}
			$\hat{\mathcal{U}}=\{(\hat{u}_{b,m},q_{b});b\in(1,...,B),m\in(1,...,M)\}$	\tcp*{augmented unlabeled mini-batch}
	        $\mathcal{L}_\mathcal{X},\mathcal{L}_\mathcal{U}=\mathrm{MixMatch}(\hat{\mathcal{X}},\hat{\mathcal{U}})$ \tcp*{apply MixMatch}
			$\mathcal{L}=\mathcal{L}_\mathcal{X}+\lambda_u \mathcal{L}_\mathcal{U}+\lambda_r \mathcal{L}_\mathrm{reg}$ \tcp*{total loss}
			$\theta^{(k)}=\mathrm{SGD}(\mathcal{L},\theta^{(k)})$  \tcp*{update model parameters}
		}
	}
	}
	\caption{\small DivideMix. Line 4-8: co-divide; Line 17-18: label co-refinement; Line 20: label co-guessing.}
	\label{alg:dividemix}
	
\end{algorithm}

In this section,
we introduce $\mathrm{DivideMix}$,
our proposed method for learning with noisy labels.
An overview of the method is shown in Figure~\ref{fig:framework}.
To avoid confirmation bias of self-training where the model would accumulate its errors,
we simultaneously train two networks to filter errors for each other through epoch-level implicit teaching and batch-level explicit teaching.
At each epoch, 
we perform co-divide,
where one network divides the noisy training dataset into a clean labeled set ($\mathcal{X}$) and a noisy unlabeled set ($\mathcal{U}$),
which are then used by the other network.
At each mini-batch,
one network utilizes both labeled and unlabeled samples to perform semi-supervised learning guided by the other network.
Algorithm~\ref{alg:dividemix} delineates the full algorithm.

\subsection{Co-Divide by Loss Modeling}

Deep networks tend to learn clean samples faster than noisy samples~\citep{Arpit_ICML_2017},
leading to lower loss for clean samples~\citep{co-teaching,Chen_ICML_2019}.
Following~\cite{Arazo_ICML_2019},
we aim to find the probability of a sample being clean by fitting a mixture model to the per-sample loss distribution.
Formally,
let {\small $\mathcal{D}=(\mathcal{X},\mathcal{Y})=\{(x_i,y_i)\}_{i=1}^N$} denote the training data,
where $x_i$ is an image and {\small $y_i \in \{0,1\}^C$} is the one-hot label over $C$ classes.
Given a model with parameters $\theta$,
the cross-entropy loss $\ell(\theta)$ reflects how well the model fits the training samples:
\begin{equation}
\ell(\theta)=\big\{\ell_i\big\}_{i=1}^N=\big\{-\sum_{c=1}^C  y_i^c \log(\mathrm{p_{model}^c}(x_i;\theta))\big\}_{i=1}^N,
\end{equation} 
where {\small $\mathrm{p^c_{model}}$} is the model's output softmax probability for class $c$.

\cite{Arazo_ICML_2019} fit a two-component Beta Mixture Model (BMM) to the max-normalized loss $\ell$ to  model the distribution of clean and noisy samples.
However, we find that BMM tends to produce undesirable flat distributions and fails when the label noise is asymmetric.
Instead,
Gaussian Mixture Model (GMM)~\citep{GMM} can better distinguish clean and noisy samples due to its flexibility in the sharpness of distribution.
Therefore,
we fit a two-component GMM to $\ell$ using the Expectation-Maximization algorithm.
For each sample,
its clean probability $w_i$ is the posterior probability $p(g|\ell_i)$,
where $g$ is the Gaussian component with smaller mean (smaller loss).

We divide the training data into a labeled set and an unlabeled set by setting a threshold $\tau$ on $w_i$.
However, training a model using the data divided by itself could lead to confirmation bias (\ie~the model is prone to confirm its mistakes~\citep{Tarvainen_NIPS_17}),
as noisy samples that are wrongly grouped into the labeled set would keep having lower loss due to the model overfitting to their labels.
Therefore,
we propose co-divide to avoid error accumulation.
In co-divide,
the GMM for one network is used to divide training data for the other network.
The two networks are kept diverged from each other due to different (random) parameter initialization,
different training data division, different (random) mini-batch sequence, and different training targets.
Being diverged offers the two networks distinct abilities to filter different types of error,
making the model more robust to noise.

\begin{figure*}[!t]
 \centering
\small
 \begin{minipage}{0.328\textwidth}
	\centering
	\includegraphics[width=\textwidth]{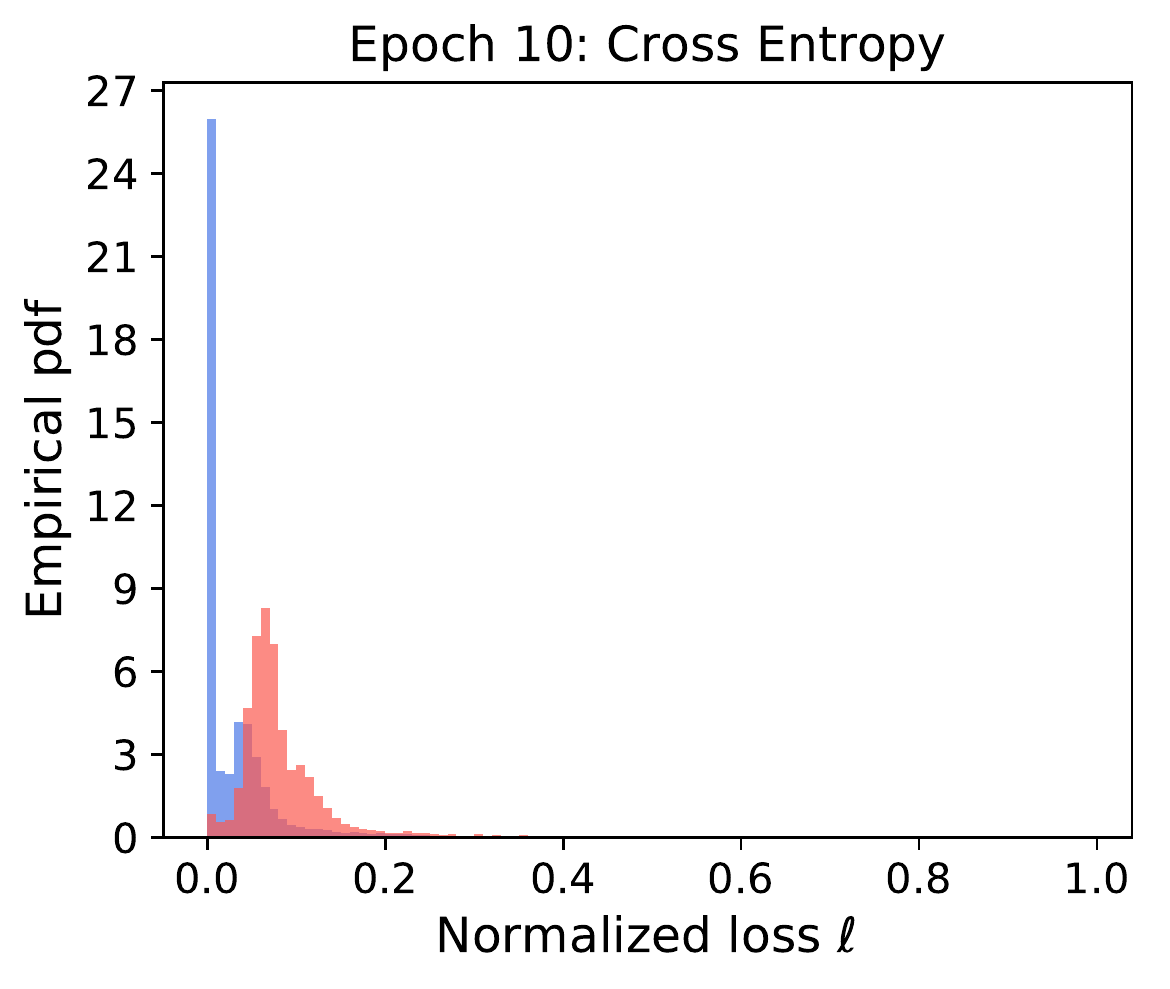}
	(a)	
\end{minipage}
 \begin{minipage}{0.328\textwidth}
	\centering
	\includegraphics[width=\textwidth]{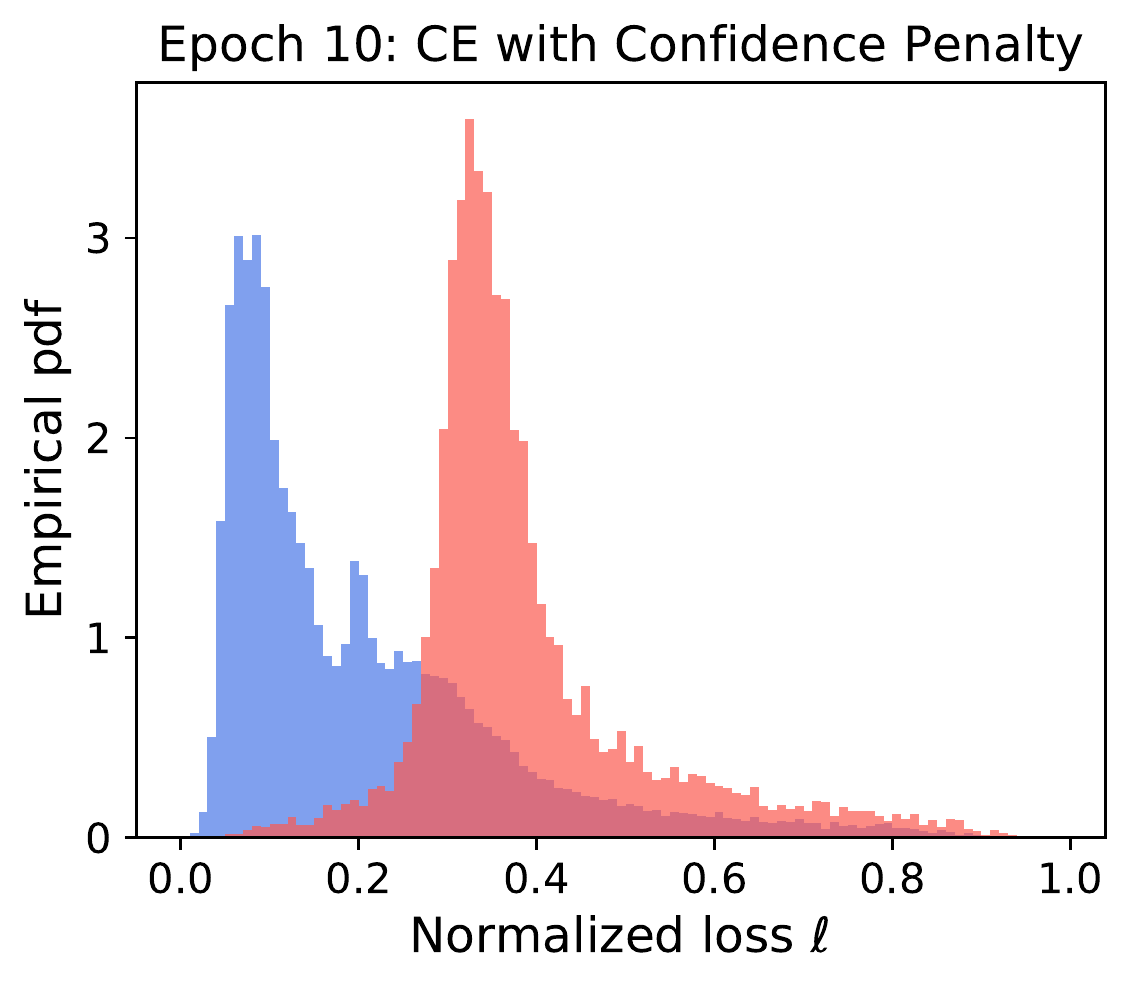}	
	(b)	
\end{minipage}
 \begin{minipage}{0.328\textwidth}
	\centering
	\includegraphics[width=\textwidth]{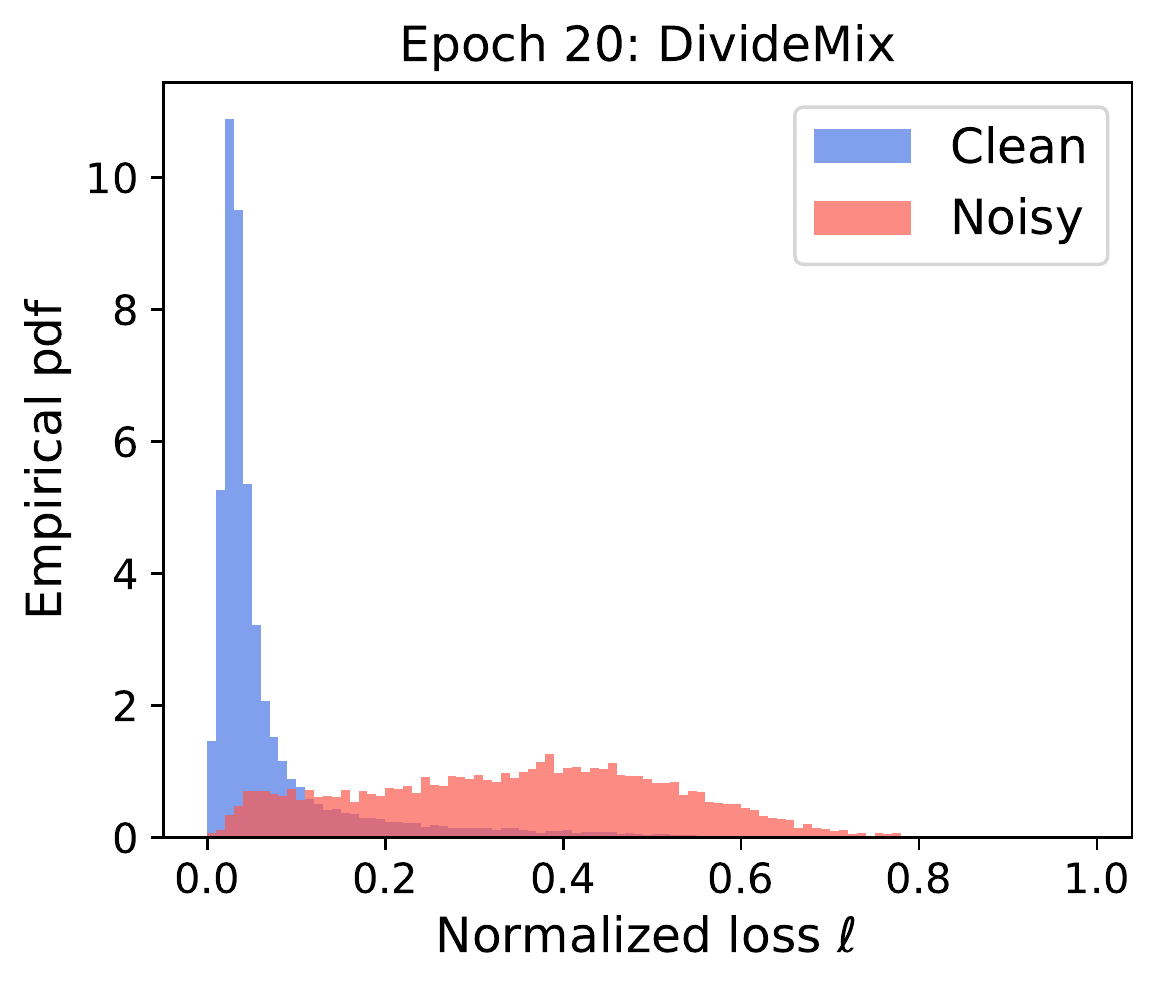}	
	 (c)	
\end{minipage}
\vspace{-1.5ex}
  \caption
  	{
  	\small
		Training on CIFAR-10 with 40\% asymmetric noise, warm up for 10 epochs. (a) Standard training with cross-entropy loss causes the model to overfit and produce over-confident predictions, making $\ell$ difficult to be modeled by the GMM. (b) Adding a confidence penalty (negative entropy) during warm up leads to more evenly-distributed $\ell$. (c) Training with $\mathrm{DivideMix}$ can effectively reduce the loss for clean samples while keeping the loss larger for most noisy samples. 
	  } 
 
  \label{fig:loss}
 \end{figure*}

\noindent\textbf{Confidence Penalty for Asymmetric Noise.}
For initial convergence of the algorithm,
we need to ``warm up" the model for a few epochs by training on all data using the standard cross-entropy loss.
The warm up is effective for symmetric (\ie~uniformly random) label noise.
However, for asymmetric (\ie~class-conditional) label noise,
the network would quickly overfit to noise during warm up and produce over-confident (low entropy) predictions,
which leads to most samples having near-zero normalized loss (see Figure~\ref{fig:loss}a).
In such cases, the GMM cannot effectively distinguish clean and noisy samples based on the loss distribution. 
To address this issue,
we penalize confident predictions from the network by adding a negative entropy term, $-\mathcal{H}$~\citep{confidence}, to the cross-entropy loss during warm up.
The entropy of a model's prediction for an input $x$ is defined as:
\begin{equation}
\label{eqn:penalty}
\mathcal{H} = -\sum_{c} \mathrm{p^c_{model}}(x;\theta) \log(\mathrm{p^c_{model}}(x;\theta)),
\end{equation}
By maximizing the entropy,
$\ell$  becomes more evenly distributed (see Figure~\ref{fig:loss}b) and easier to be modeled by the GMM.
Furthermore,
in Figure~\ref{fig:loss}c we show $\ell$ when the model is trained with $\mathrm{DivideMix}$ for 10 more epochs after warm up.
The proposed method can significantly reduce the loss for clean samples while keeping the loss larger for most noisy samples.

\subsection{MixMatch with Label Co-Refinement and Co-Guessing}
At each epoch,
having divided the training data,
we train the two networks one at a time while keeping the other one fixed.
Given a mini-batch of labeled samples with their corresponding one-hot labels and clean probability, $\{(x_b,y_b,w_b);b\in(1,...,B)\}$,
and a mini-batch of unlabeled samples $\{u_b;b\in(1,...,B)\}$,
we exploit $\mathrm{MixMatch}$~\citep{mixmatch} for SSL.
$\mathrm{MixMatch}$ utilizes unlabeled data by merging consistency regularization (\ie~encourage the model to output same predictions on perturbed unlabeled data) and entropy minimization (\ie~encourage the model to output confident predictions on unlabeled data) with the  $\mathrm{MixUp}$~\citep{mixup} augmentation (\ie~encourage the model to have linear behaviour between samples).

To account for label noise,
we make two improvements to $\mathrm{MixMatch}$ which enable the two networks to teach each other. 
First, we perform label co-refinement for labeled samples by linearly combining the ground-truth label $y_b$ with the network's prediction $p_b$ (averaged across multiple augmentations of $x_b$),
guided by the clean probability $w_b$ produced by the other network:
\begin{equation}
\bar{y}_b=w_b y_b+(1-w_b){p}_{b}.
\end{equation}

Then we apply a sharpening function on the refined label to reduce its temperature:
\begin{equation}
\small
\hat{y}_b=\mathrm{Sharpen}(\bar{y}_b,T)={\bar y_b^c}^{\frac{1}{T}} \bigg/ \sum_{c=1}^C {\bar y_b^c}^{\frac{1}{T}},~\text{for}~ c=1,2,...,C.
\end{equation}
Second, we use the ensemble of predictions from both networks to ``co-guess'' the labels for unlabeled samples (algorithm~\ref{alg:dividemix}, line 20),
which can produce more reliable guessed labels.

Having acquired $\hat{\mathcal{X}}$ (and $\hat{\mathcal{U}}$) which consists of multiple augmentations of labeled (unlabeled) samples and their refined (guessed) labels,
we follow $\mathrm{MixMatch}$ to ``mix'' the data,
where each sample is interpolated with another sample randomly chosen from the combined mini-batch of  $\hat{\mathcal{X}}$ and $\hat{\mathcal{U}}$.
Specifically,
for a pair of samples $(x_1,x_2)$ and their corresponding labels $(p_1,p_2)$,
the mixed $(x',p')$ is computed by:
{\small
\begin{align}
\lambda &\sim \mathrm{Beta}(\alpha,\alpha),\\
\lambda' &= \max(\lambda,1-\lambda), \label{eqn:max}\\
x'&=\lambda' x_1+(1-\lambda')x_2,\\
p'&=\lambda' p_1+(1-\lambda')p_2.
\end{align}
}%

$\mathrm{MixMatch}$ transforms $\hat{\mathcal{X}}$ and $\hat{\mathcal{U}}$ into $\mathcal{X}'$ and $\mathcal{U}'$.
Equation~\ref{eqn:max} ensures that $\mathcal{X}'$ are ``closer'' to $\hat{\mathcal{X}}$ than $\hat{\mathcal{U}}$.
The loss on $\mathcal{X}'$ is the cross-entropy loss and the loss on $\mathcal{U}'$ is the mean squared error:
{\small
\begin{align}
\mathcal{L}_\mathcal{X} &= -\frac{1}{|\mathcal{X}'|} \sum_{x,p\in\mathcal{X}'} \sum_{c} p_c \log(\mathrm{p_{model}^c}(x;\theta)),\\
\mathcal{L}_\mathcal{U} &= \frac{1}{|\mathcal{U}'|}  \sum_{x,p\in\mathcal{U}'} \left\lVert {p-\mathrm{p_{model}}(x;\theta)} \right\rVert^2_2.
\end{align}
}%

Under high levels of noise,
the network would be encouraged to predict the same class to minimize the loss.
To prevent assigning all samples to a single class,
we apply the regularization term used by~\cite{Tanaka_CVPR_2018} and~\cite{Arazo_ICML_2019},
which uses a uniform prior distribution $\pi$ (\ie~$\pi_c=1 \small/ C$) to regularize the model's average output across all samples in the mini-batch:
\begin{equation}
\small
\mathcal{L}_\mathrm{reg}= \sum_c \pi_c \log({\pi_c}\bigg/ \frac{1}{|\mathcal{X}'|+|\mathcal{U}'|} \sum_{x \in\mathcal{X}'+\mathcal{U}'} \mathrm{p_{model}^c}(x;\theta)).
\end{equation}

Finally, the total loss is:
\begin{equation}
\mathcal{L}=\mathcal{L}_\mathcal{X}+\lambda_u \mathcal{L}_\mathcal{U}+\lambda_r \mathcal{L}_\mathrm{reg}.
\end{equation}
In our experiments, we set $\lambda_r$ as 1 and use $\lambda_u$ to control the strength of the unsupervised loss.
\section{Experiments}
\label{sec:experiment}
\subsection{Datasets and Implementation Details}

We extensively validate our method on four benchmark datasets,
namely CIFAR-10, CIFAR-100~\citep{cifar}, Clothing1M~\citep{Tong_CVPR_2015},
and WebVision~\citep{webvision}.
Both CIFAR-10 and CIFAR-100 contain 50K training images and 10K test images of size $32\times32$.
Following previous works~\citep{Tanaka_CVPR_2018,MLNT},
we experiment with two types of label noise: \textit{symmetric} and \textit{asymmetric}. 
Symmetric noise is generated by randomly replacing the labels for a percentage of the training data with all possible labels.
Note that there is another criterion for symmetric label noise injection where the true labels cannot be maintained ~\citep{Jiang_ICML_2018,Wang_CVPR_2018},
for which we also report the results (Table~\ref{tbl:cifar_sn} in Appendix).
Asymmetric noise is designed to mimic the structure of real-world label noise,
where labels are only replaced by similar classes~(\eg~deer$\rightarrow$horse, dog$\leftrightarrow$cat).

We use an 18-layer PreAct Resnet~\citep{He_ECCV_2016} and train it using SGD with a momentum of 0.9, 
a weight decay of 0.0005, 
and a batch size of 128.
The network is trained for 300 epochs.
We set the initial learning rate as 0.02, 
and reduce it by a factor of 10 after 150 epochs.
The warm up period is 10 epochs for CIFAR-10 and 30 epochs for CIFAR-100.
We find that most hyperparameters introduced by $\mathrm{DivideMix}$ do not need to be heavily tuned.
For all CIFAR experiments,
we use the same hyperparameters $M=2$, $T=0.5$, and $\alpha=4$.
$\tau$ is set as 0.5 except for $90\%$ noise ratio when it is set as 0.6.
We choose $\lambda_u$ from $\{0,25,50,150\}$ using a small validation set.

Clothing1M and WebVision 1.0 are two large-scale datasets with real-world noisy labels.
Clothing1M consists of 1 million training images collected from online shopping websites
with labels generated from surrounding texts. 
We follow previous work~\citep{MLNT} and use ResNet-50 with ImageNet pretrained weights.
WebVision contains 2.4 million images crawled from the web using the 1,000 concepts in ImageNet ILSVRC12.
Following previous work~\citep{Chen_ICML_2019}, we compare baseline methods on the first 50 classes of the Google image subset using the inception-resnet v2~\citep{InceptionResnet}.
The training details are delineated in Appendix B.

\subsection{Comparison with State-of-the-art Methods}

We compare $\mathrm{DivideMix}$ with multiple baselines using the same network architecture.
Here we introduce some of the most recent state-of-the-art methods:
Meta-Learning~\citep{MLNT} proposes a gradient based method to find model parameters that are more noise-tolerant;
Joint-Optim~\citep{Tanaka_CVPR_2018} and P-correction~\citep{Yi_2019_CVPR} jointly optimize the sample labels and the network parameters;
M-correction~\citep{Arazo_ICML_2019} models sample loss with BMM and applies $\mathrm{MixUp}$.
Note that none of these methods can consistently outperform others across different datasets.
M-correction excels at symmetric noise,
whereas Meta-Learning performs better for asymmetric noise.

\begin{table}[!t]
	\centering
	\small
	\begin{tabular}{l l |c|c|c|c||c|c|c|c} 
		\toprule	 	
		   Dataset & &\multicolumn{4}{c||}{CIFAR-10}&\multicolumn{4}{c}{CIFAR-100}\\
			\midrule 
		   Method/Noise ratio& & 20\% & 50\%& 80\% & 90\% &20\% & 50\%& 80\% & 90\% \\
			\midrule
			\multirow{2}{*}{Cross-Entropy} & Best &  86.8 &  79.4& 62.9 & 42.7    &62.0& 46.7&19.9 &10.1\\
			&Last&82.7& 57.9& 26.1&16.8    &61.8& 37.3& 8.8&3.5 \\
			\midrule
			Bootstrap & Best &  86.8&79.8 &63.3 &42.9    &62.1 &46.6 &19.9 &10.2\\
			\citep{Reed_2015_ICLR}&Last&82.9& 58.4&26.8 & 17.0   & 62.0&37.9 &8.9 &3.8\\
			\midrule
			F-correction& Best &  86.8 &79.8 & 63.3& 42.9   &61.5 &46.6 &19.9 &10.2\\
			\citep{Giorgio_CVPR_2017}&Last&83.1&59.4 &26.2 & 18.8   & 61.4& 37.3& 9.0&3.4\\
			\midrule	
			Co-teaching$+^*$& Best & 89.5 &85.7 &67.4 &47.9     &  65.6&51.8 &27.9 & 13.7    \\
			\citep{Yu_ICML_2019}&Last& 88.2 & 84.1& 45.5& 30.1  & 64.1 & 45.3& 15.5&   8.8  \\
			\midrule					
			Mixup & Best &  95.6&87.1 &71.6 &52.2    & 67.8& 57.3&30.8 &14.6\\
			\citep{mixup}&Last&92.3& 77.6&46.7 &43.9    & 66.0&46.6 &17.6 &8.1\\	
			\midrule									
			P-correction$^*$& Best &92.4  & 89.1&77.5 & 58.9    &69.4  & 57.5& 31.1&  15.3   \\
			\citep{Yi_2019_CVPR}&Last& 92.0 & 88.7&76.5 &  58.2   &  68.1& 56.4&20.7 &   8.8  \\			
			\midrule
			Meta-Learning$^*$& Best & 92.9 & 89.3&77.4 &  58.7   & 68.5 &59.2  &42.4 &19.5\\
			\citep{MLNT}&Last& 92.0 & 88.8& 76.1& 58.3    &  67.7&  58.0& 40.1&14.3\\
			\midrule				
			M-correction & Best &  94.0& 92.0& 86.8&69.1 	& 73.9&66.1 &48.2 &24.3\\
			\citep{Arazo_ICML_2019}&Last&93.8&91.9 &86.6 & 68.7	& 73.4&65.4 &47.6 &20.5	\\
			\midrule		
			\multirow{2}{*}{DivideMix}	& Best &\textbf{96.1}&\textbf{94.6}	&\textbf{93.2}&\textbf{76.0} & \textbf{77.3}  & \textbf{74.6}&\textbf{60.2}  &\textbf{31.5} \\	            
			& Last & \textbf{95.7} &\textbf{94.4}  & \textbf{92.9} & \textbf{75.4}    &  \textbf{76.9} & \textbf{74.2}& \textbf{59.6} & \textbf{31.0}\\	 						            
		\bottomrule
\end{tabular}
	\caption
		{
			Comparison with state-of-the-art methods in test accuracy (\%) on CIFAR-10 and CIFAR-100 with symmetric noise. Methods marked by * denote re-implementations based on public code.
		}
	\label{tbl:cifar}
\end{table}

Table~\ref{tbl:cifar} shows the results on CIFAR-10 and CIFAR-100 with different levels of symmetric label noise ranging from $20\%$ to $90\%$.
We report both the best test accuracy across all epochs and the averaged test accuracy over the last 10 epochs.
$\mathrm{DivideMix}$ outperforms state-of-the-art methods by a large margin across all noise ratios.
The improvement is substantial ({\small$\sim$}10\% in accuracy) for the more challenging CIFAR-100 with high noise ratios. Appendix A shows comparison with more methods in Table~\ref{tbl:cifar_sn}. 
The results on CIFAR-10 with asymmetric noise is shown in Table~\ref{tbl:cifar_an}.
We use $40\%$ because certain classes become theoretically indistinguishable for asymmetric noise larger than $50\%$.

\begin{table}[ht]
	\centering
    \small
	\begin{tabular}	{l |c|c}
		\toprule	 	
			Method & Best &  Last\\
			\midrule			
			Cross-Entropy & 85.0&72.3\\			
			F-correction~\citep{Giorgio_CVPR_2017} & 87.2&83.1\\
			M-correction~\citep{Arazo_ICML_2019}& 87.4 & 86.3\\		
			Iterative-CV~\citep{Chen_ICML_2019}& 88.6&88.0\\	
			P-correction~\citep{Yi_2019_CVPR}& 88.5&88.1\\					
			Joint-Optim~\citep{Tanaka_CVPR_2018} & 88.9&88.4\\						
			Meta-Learning~\citep{MLNT} & 89.2&88.6\\						
			\midrule
			DivideMix& \textbf{93.4} & \textbf{92.1}\\
		\bottomrule
	\end{tabular}
	\caption
		{
			Comparison with state-of-the-art methods in test accuracy (\%) on CIFAR-10 with 40\% asymmetric noise. 
			We re-implement all methods under the same setting.
		}
		\label{tbl:cifar_an}
\end{table}		

Table~\ref{tbl:clothing} and Table~\ref{tbl:webvision} show the results on Clothing1M and WebVision, respectively.
$\mathrm{DivideMix}$ consistently outperforms state-of-the-art methods across all datasets with different types of label noise. For WebVision, we achieve more than 12\% improvement in top-1 accuracy.

\begin{table}[!t]
	\centering
	\small
	\begin{tabular}	{l |c }
		\toprule	 	
			Method  & Test Accuracy \\
			\midrule			
			Cross-Entropy & 69.21 \\
			F-correction~\citep{Giorgio_CVPR_2017}  &69.84\\		
			M-correction~\citep{Arazo_ICML_2019}&  71.00 \\			
			Joint-Optim~\citep{Tanaka_CVPR_2018}  & 72.16\\			
			Meta-Cleaner~\citep{MetaCleaner} & 72.50\\
			Meta-Learning~\citep{MLNT}  & 73.47\\				
			P-correction~\citep{Yi_2019_CVPR}&73.49\\
			\midrule
			DivideMix & \textbf{74.76}\\
		\bottomrule
	\end{tabular}
	\caption
		{
		Comparison with state-of-the-art methods in test accuracy (\%) on Clothing1M. Results for baselines are copied from original papers.
		}
	\label{tbl:clothing}	
\end{table}		
\begin{table}[ht]
	\centering
	\small
	\begin{tabular}	{l |c|c|c|c }
		\toprule	 	
			\multirow{2}{*}{Method}  & \multicolumn{2}{c|}{WebVision} & \multicolumn{2}{c}{ILSVRC12}\\
			\cmidrule{2-5}
			& top1 & top5& top1 & top5\\
			\midrule			
			F-correction~\citep{Giorgio_CVPR_2017} & 61.12 & 82.68&  57.36 &82.36\\		
		    
			Decoupling~\citep{decouple} & 62.54 &84.74& 58.26 &82.26\\
				D2L~\citep{Ma_ICML_2018} &62.68 &84.00&  57.80 &81.36\\
			MentorNet~\citep{Jiang_ICML_2018}  &63.00 &81.40&  57.80 &79.92\\	
			
			Co-teaching~\citep{co-teaching}&63.58 &85.20&   61.48 &84.70\\			
			Iterative-CV~\citep{Chen_ICML_2019} 	&  65.24 &85.34&  61.60 &84.98\\	
			\midrule
			DivideMix &\textbf{77.32} &\textbf{91.64}& \textbf{75.20} &\textbf{90.84}\\
		\bottomrule
	\end{tabular}
	\caption
		{
		Comparison with state-of-the-art methods trained on (mini) WebVision dataset. Numbers denote top-1 (top-5) accuracy (\%) on the WebVision validation set and the ImageNet ILSVRC12 validation set.
		Results for baseline methods are copied from~\cite{Chen_ICML_2019}.
		}
	\label{tbl:webvision}

\end{table}

\subsection{Ablation Study}

We study the effect of removing different components to provide insights into what makes $\mathrm{DivideMix}$ successful. 
We analyze the results in Table~\ref{tbl:ablation} as follows.
Appendix C contains additional explanations.

\begin{table}[h]
	\centering
	\renewcommand{\arraystretch}{0.9}
	\resizebox{\textwidth}{!}{  
	\begin{tabular}{l l | c |c|c|c|c||c|c|c|c} 
		\toprule	 	
			Dataset & &\multicolumn{5}{c||}{CIFAR-10}&\multicolumn{4}{c}{CIFAR-100}\\
			\midrule
			Noise type&  & \multicolumn{4}{c|}{Sym.}& Asym. & \multicolumn{4}{c}{Sym.}\\
			\midrule
			Methods/Noise ratio&  & 20\% & 50\%& 80\% & 90\% &40\%&20\% & 50\%& 80\% & 90\% \\
			\midrule			
			\multirow{2}{*}{DivideMix}	& Best &\textbf{96.1}&\textbf{94.6}	&\textbf{93.2}&\textbf{76.0}&\textbf{93.4} & \textbf{77.3}  & \textbf{74.6}&\textbf{60.2}  &\textbf{31.5} \\	            
			& Last  & \textbf{95.7} &\textbf{94.4}  & \textbf{92.9} & \textbf{75.4} &\textbf{92.1}  &  \textbf{76.9} & \textbf{74.2}& \textbf{59.6} & \textbf{31.0}\\	 						 	
			\midrule
			\multirow{2}{*}{DivideMix with $\theta^{(1)}$ test}	& Best&95.2 & 94.2& 93.0& 75.5& 92.7  & 75.2& 72.8& 58.3& 29.9\\	            
			& Last &95.0 & 93.7&92.4 &74.2&91.4   &74.8& 72.1& 57.6& 29.2\\	            
			\midrule
			\multirow{2}{*}{DivideMix w/o co-training}	& Best & 95.0& 94.0&92.6 & 74.3& 91.9& 74.8&72.3 &56.7 & 27.7\\	            
			& Last &  94.8&93.3 &92.2 &73.2&90.6 &74.1 &71.7 & 56.3&27.2 \\	            
			\midrule			
			\multirow{2}{*}{DivideMix w/o label refinement}	& Best  &96.0 & 94.6& 93.0&73.7& 87.7& 76.9&74.2&  58.7& 26.9\\	                    
			& Last &95.5 & 94.2& 92.7& 73.0&86.3 &76.4 & 73.9&58.2 & 26.3\\	            
			\midrule
			\multirow{2}{*}{DivideMix w/o augmentation}& Best &95.3 & 94.1&92.2 &73.9&89.5   & 76.5& 73.1& 58.2& 26.9\\	            
			& Last& 94.9&93.5 &91.8 & 73.0& 88.4& 76.2& 72.6& 58.0&26.4 \\	            
			\midrule
			\multirow{2}{*}{Divide and $\mathrm{MixMatch}$}& Best &94.1 & 92.8&89.7&70.1 &86.5  & 73.7& 70.5&55.3 &25.0 \\	            
		    & Last& 93.5&92.3 &89.1 & 68.6& 85.2& 72.4&69.7 &53.9&23.7 \\	            			
		\bottomrule
\end{tabular}
}

	\caption
		{
			Ablation study results in terms of test accuracy (\%) on CIFAR-10 and CIFAR-100.
		}
	\label{tbl:ablation}
\end{table}

\vspace{-\topsep}
\begin{itemize}[leftmargin=*]
	\setlength\itemsep{0pt}
	\item To study the effect of model ensemble during \textit{test},
	we use the prediction from a single model $\theta^{(1)}$ instead of averaging the predictions from both networks as in $\mathrm{DivideMix}$.
	Note that the training process remains unchanged. 
	The decrease in accuracy suggests that the ensemble of two diverged networks consistently yields better  performance during inference.
	\item
	To study the effect of co-training, 
	we train a single network using self-divide (\ie~divide the training data based on its own loss).
	The performance further decreases compared to $\theta^{(1)}$. 
	\item 
	We find that both label refinement and input augmentation are beneficial for $\mathrm{DivideMix}$.
	\item 
	We combine self-divide with the original $\mathrm{MixMatch}$ as a naive baseline for using SLL in LNL.
\end{itemize}
\vspace{-\topsep}

Appendix A also introduces more in-depth studies in examining the robustness of our method to label noise, including the AUC for clean/noisy sample classification on CIFAR-10 training data, qualitative examples from Clothing1M where our method can effectively identify the noisy samples and leverage them as unlabeled data, and visualization results using t-SNE.

\section{Conclusion}
\label{sec:conclusion}
In this paper, we propose $\mathrm{DivideMix}$ for learning with noisy labels by leveraging SSL. Our method trains two networks simultaneously and achieves robustness to noise through dataset co-divide, label co-refinement and co-guessing.
Through extensive experiments across multiple datasets, we show that $\mathrm{DivideMix}$ consistently exhibits substantial performance improvements compared to state-of-the-art methods. For future work, we are interested in incorporating additional ideas from SSL to LNL, and vice versa. Furthermore, we are also interested in adapting $\mathrm{DivideMix}$ to other domains such as NLP.


\bibliography{reference}
\bibliographystyle{iclr2020_conference}

\newpage
\begin{appendices}

\section{\normalsize Additional Experiment Results}

In Table~\ref{tbl:cifar_sn}, we compare $\mathrm{DivideMix}$ with previous methods under the symmetric noise setting where true labels cannot be maintained. $\mathrm{DivideMix}$ significantly outperforms previous methods which use deeper or wider network architectures.

\begin{table}[ht]
	\centering
	\setlength\tabcolsep{4.5pt}
	\resizebox{\textwidth}{!}{  
	\begin{tabular}{l | l|c|c|c|c||c|c|c|c} 
		\toprule	 	
			\multirow{2}{*}{Method} & 	\multirow{2}{*}{Architecture} &\multicolumn{4}{c||}{CIFAR-10}&\multicolumn{4}{c}{CIFAR-100}\\
			\cmidrule{3-10}
		    & &20\% & 40\%& 60\% & 80\% &20\% & 40\%& 60\% & 80\% \\
			\midrule
			MentorNet~\citep{Jiang_ICML_2018} &WRN-101 &92.0&  89.0& - & 49.0   &73.0& 68.0&- &35.0\\
			\midrule
			D2L~\citep{Ma_ICML_2018} &CNN-12/RN-44 &85.1&  83.4& 72.8 & -   &62.2& 52.0& 42.3 &-\\
			\midrule
			Reweight~\citep{Ren_ICML_2018} &WRN-28 & 86.9 & - & - & - & 61.3 & - & - & -\\
			\midrule
			Abstention~\citep{Abstention}&WRN-28&93.4 & 90.9 & 87.6 & 70.8 & 75.8 & 68.2 & 59.4 & 34.1\\
			\midrule
			DivideMix& PRN-18&\textbf{96.2} & \textbf{94.9}& \textbf{94.3}& \textbf{79.8}& \textbf{77.2}&\textbf{75.2}&\textbf{72.0}&\textbf{60.0}\\
				            
		\bottomrule
	
\end{tabular}
}
	\caption
		{
			Comparison with state-of-the-art methods in test accuracy (\%) on CIFAR-10 and CIFAR-100 with symmetric noise. Numbers are copied from original papers.
			Key: WRN (Wide ResNet), PRN (PreActivation ResNet).
			$\mathrm{DivideMix}$ outperforms previous methods that use deeper/wider networks.
		}
	\label{tbl:cifar_sn}
\end{table}

\begin{figure*}[ht]
 \centering
  	\includegraphics[width=0.65\textwidth]{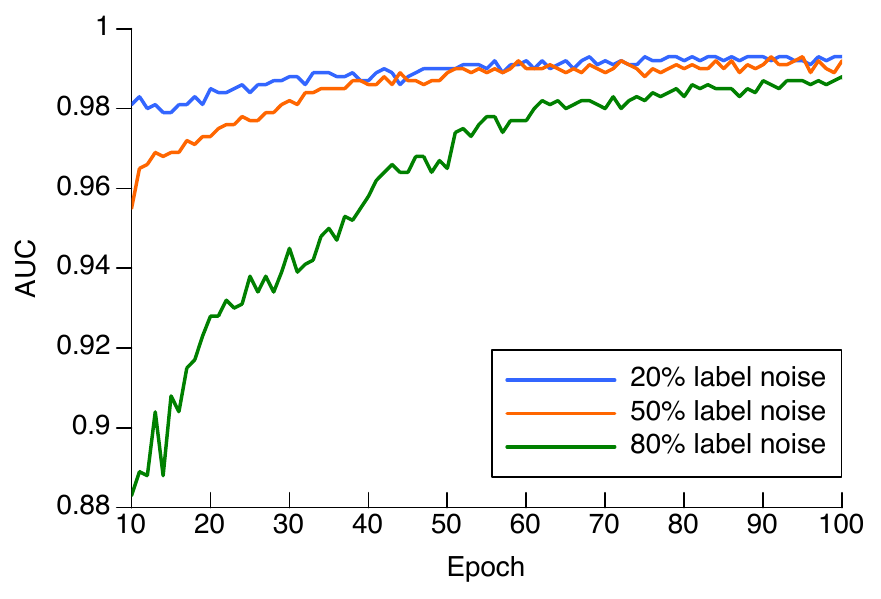}
  	 \vspace{-2ex}
  \caption
  	{
  	\small
		Area Under a Curve for clean/noisy image classification on CIFAR-10 training samples. Our method can effectively filter out the noisy samples and leverage them as unlabeled data.
    } 
  \label{fig:auc}
 \end{figure*}

In Figure~\ref{fig:auc}, we show the Area Under a Curve (AUC) for clean/noisy sample classification on CIFAR-10 training data from one of the GMMs during the first 100 epochs. Our method can effectively separate clean and noisy samples as training proceeds, even for high noise ratio.

\begin{figure*}[ht]
 \centering
  	\includegraphics[width=1.0\textwidth]{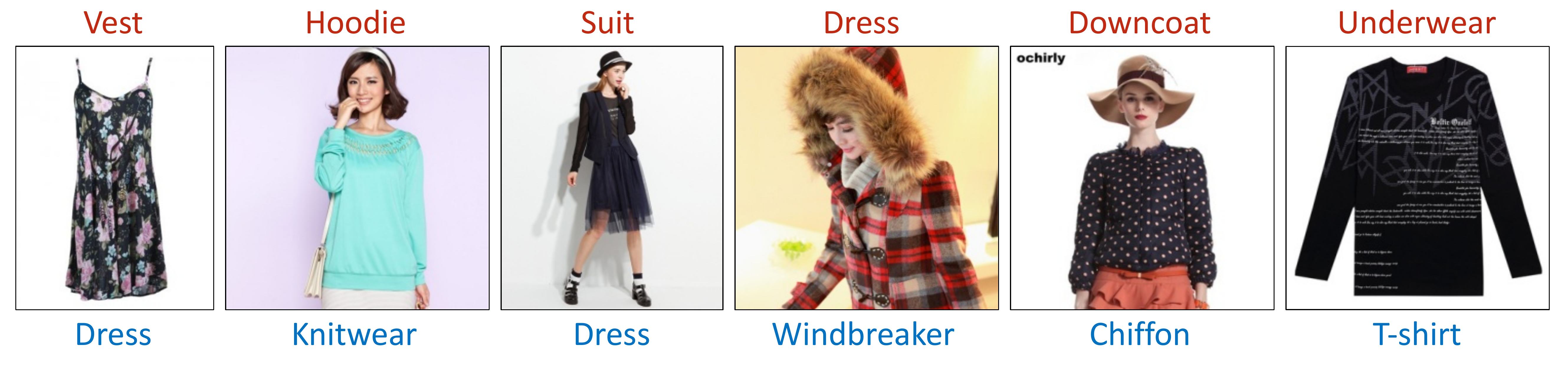}
  	 \vspace{-5ex}
  \caption
  	{
  	\small
  	Clothing1M images identified as noisy samples by our method. Ground-truth labels are shown above in red and the co-guessed labels are shown below in blue. 
	  } 

  \label{fig:clothing_example}
 \end{figure*}

In Figure~\ref{fig:clothing_example},
we show example images in Clothing1M identified by our method as noisy samples.
Our method achieves noise filtering by discarding the noisy labels (shown in red) and using the co-guessed labels (shown in blue) to regularize training.

\begin{figure*}[ht]
 \centering
\small
 \begin{minipage}{0.45\textwidth}
	\centering
	\includegraphics[width=\textwidth]{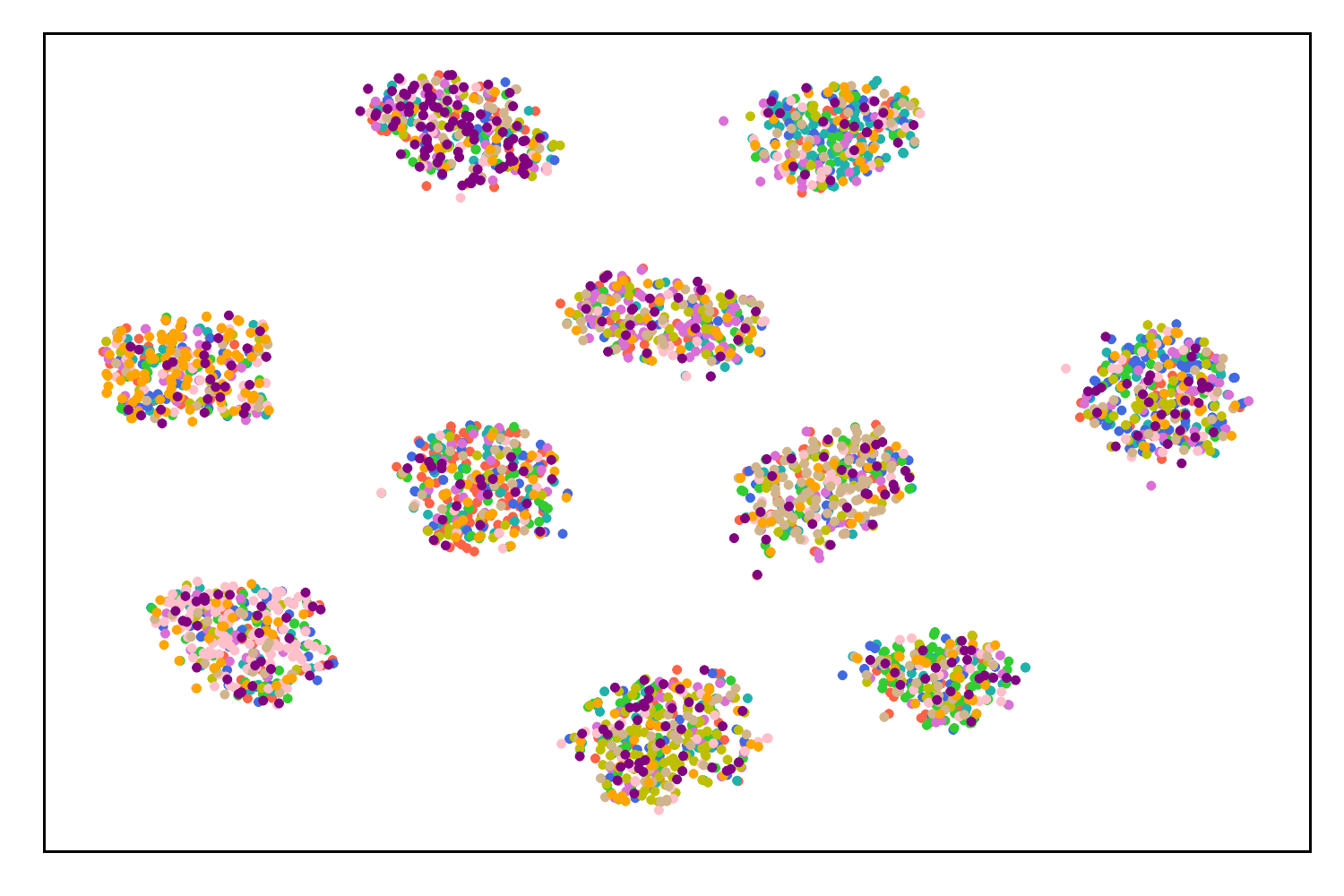}
	(a)	Noisy training labels.	
\end{minipage}
\hspace{2ex}
 \begin{minipage}{0.45\textwidth}
	\centering
	\includegraphics[width=\textwidth]{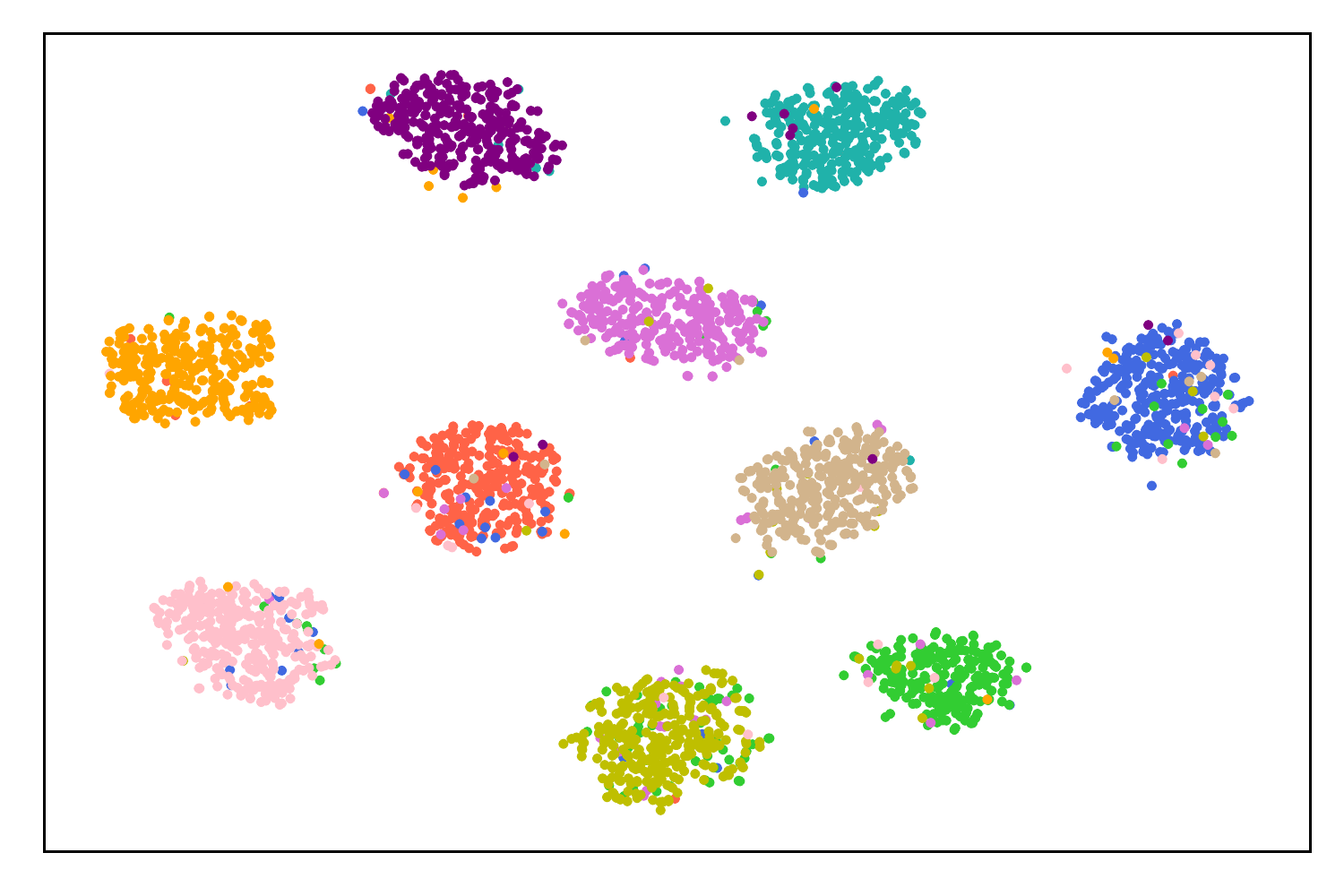}	
	(b) True labels.
\end{minipage}

  \caption
  	{
  	\small
		T-SNE of training images after training the model using $\mathrm{DivideMix}$ for 200 epochs on CIFAR-10 with 80\% label noise. Different colors indicate (a) noisy training labels or (b) true labels.
		$\mathrm{DivideMix}$ is able to learn the true class distribution of the training data despite the label noise.
	  } 
 
  \label{fig:tsne}
 \end{figure*}

In Figure~\ref{fig:tsne}, we visualize the features of training images using t-SNE~\citep{tsne}.
The model is trained using $\mathrm{DivideMix}$ for 200 epochs on CIFAR-10 with 80\% label noise. The embeddings form 10 distinct clusters corresponding to the true class labels,
not the noisy training labels, which demonstrates our method's robustness to label noise.

\section{\normalsize Additional Training Details}

For CIFAR experiments, the only hyperparameter that we tune on a per-experiment basis is the unsupervised loss weight $\lambda_u$. Table~\ref{tbl:params_cifar} shows the value that we use.
A larger $\lambda_u$ is required for stronger regularization under high noise ratios or with more classes. 

For both Clothing1M and WebVision, we use the same set of hyperparameters $M=2$, $T=0.5$, $\tau=0.5$, $\lambda_u=0$, $\alpha=0.5$, and train the network using SGD with a momentum of 0.9,  a weight decay of 0.001, 
and a batch size of 32. The warm up period is 1 epoch. For Clothing1M,
we train the network for 80 epochs. The initial learning rate is set as 0.002 and reduced by a factor of 10 after 40 epochs. For each epoch, we sample 1000 mini-batches from the training data while ensuring the labels (noisy) are balanced. For WebVision, we train the network for 100 epochs. The initial learning rate is set as 0.01 and reduced by a factor of 10 after 50 epochs.

\begin{table}[ht]
	\centering
	\small
	\begin{tabular}{c|c|c|c|c|c||c|c|c|c} 
		\toprule	 	
			\multirow{2}{*}{Hyperparameter} &\multicolumn{5}{c||}{CIFAR-10}&\multicolumn{4}{c}{CIFAR-100}\\
			\cmidrule{2-10}
			&Asym. 40\% & 20\% & 50\%& 80\% & 90\% &20\%  & 50\%& 80\% & 90\% \\
			\midrule
			$\lambda_u$  &0&0& 25& 25& 50  &25& 150& 150&150 \\				            
		\bottomrule
\end{tabular}
	\caption
		{
			Unsupervised loss weight $\lambda_u$ for CIFAR experiments. Higher noise ratio requires stronger regularization from unlabeled samples.
		}
	\label{tbl:params_cifar}
\end{table}

\section{\normalsize Additional Explanations for Ablation Study}

Here we clarify some details for the baseline methods in the ablation study.
First, \textit{DivideMix w/o co-training} still has dataset division, label refinement and label guessing,
but performed by the same model.
Thus, the performance drop (especially for CIFAR-100 with high noise ratio) suggests the disadvantage of self-training. 
Second, \textit{label refinement} is important for high noise ratio because more noisy samples would be mistakenly divided into the labeled set.
Third, \textit{augmentation} improves performance through both producing more reliable predictions and achieving consistency regularization.
In addition, same as~\cite{mixmatch}, we also find that \textit{temperature sharpening} is essential for our method to perform well.

\section{\normalsize Training Time Analysis}
We analyse the training time of $\mathrm{DivideMix}$ to understand its efficiency.
In Table~\ref{tbl:time_compare}, we compare the total training time of $\mathrm{DivideMix}$ on CIFAR-10 with several state-of-the-art methods, using a single Nvidia V100 GPU.
$\mathrm{DivideMix}$ is slower than Co-teaching$+$~\citep{Yu_ICML_2019},
but faster than P-correction~\citep{Yi_2019_CVPR} and Meta-Learning~\citep{MLNT} which involve multiple training iterations. 
In Table 9, we also break down the computation time for each operation in $\mathrm{DivideMix}$. 

\begin{table}[ht]
	\centering
	\small
	\begin{tabular}{c|c|c||c} 
		\toprule	 	
			 Co-teaching$+^*$ & P-correction &Meta-Learning& DivideMix \\
			\midrule
			4.3 h & 6.0 h & 8.6 h &  5.2 h  \\     
		\bottomrule
\end{tabular}
	\caption
		{
			Comparison of total training time (hours) on CIFAR-10.
		}
	\label{tbl:time_compare}
\end{table}

\begin{table}[ht]
	\centering
	\small
	\resizebox{\textwidth}{!}{  
	\begin{tabular}{c|c|c} 
		\toprule	 	
			 Co-Divide (Alg.~\ref{alg:dividemix}, line 4-8) & Data MixMatch (Alg.~\ref{alg:dividemix}, line 12-24) & Forward-Backward (Alg.~\ref{alg:dividemix}, line 25-27) \\
			\midrule
			 17.2 s & 16.0 s & 12.5 s\\       
		\bottomrule
\end{tabular}
}
	\caption
		{
			Computation time (seconds) per-epoch for each operation in DivideMix during training.
		}
	\label{tbl:time_component}
\end{table}

\end{appendices}

\end{document}